\documentclass[10pt,twocolumn,letterpaper]{article}

\usepackage[utf8]{inputenc}
\usepackage[T1]{fontenc}
\usepackage{geometry}
\usepackage{times}      
\usepackage{graphicx}   
\usepackage{booktabs}   

\usepackage{amsmath, amssymb, amsfonts} 
\usepackage[hyphens]{url} 
\usepackage[hidelinks]{hyperref} 
\usepackage{caption}    
\usepackage[numbers,sort&compress]{natbib} % Added sort&compress for cleaner [1-3]
\usepackage{xurl} % Better URL breaking than standard url package
\usepackage{siunitx}
\usepackage{authblk}

\title{SynthAgent: A Multi-Agent LLM Framework for Realistic Patient Simulation - A Case Study in Obesity with Mental Health Comorbidities}

\author[1]{Arman Aghaee\thanks{aaghaee@klick.com}}
\author[1]{Sepehr Asgarian\thanks{sasgarian@klick.com}}
\author[1]{Jouhyun Jeon\thanks{cjeon@klick.com}}

\affil[1]{Klick Health, Toronto, ON, Canada}

\date{}

\begin{document}

\maketitle

\begin{abstract}
Simulating high-fidelity patients offers a powerful avenue for studying complex diseases while addressing the challenges of fragmented, biased, and privacy-restricted real-world data. In this study, we introduce SynthAgent, a novel Multi-Agent System (MAS) framework designed to model obesity patients with comorbid mental disorders, including depression, anxiety, social phobia, and binge eating disorder. SynthAgent integrates clinical and medical evidence from claims data, population surveys, and patient-centered literature to construct personalized virtual patients enriched with personality traits that influence adherence, emotion regulation, and lifestyle behaviors. Through autonomous agent interactions, the system simulates disease progression, treatment response, and life management across diverse psychosocial contexts. Evaluation of more than 100 generated patients demonstrated that GPT-5 and Claude 4.5 Sonnet achieved the highest fidelity as the core engine in the proposed MAS framework, outperforming Gemini 2.5 Pro and DeepSeek-R1. SynthAgent thus provides a scalable and privacy-preserving framework for exploring patient journeys, behavioral dynamics, and decision-making processes in both medical and psychological domains.
\end{abstract}

\section{Introduction}
Obesity is a complex, multifaceted disease that involves not only metabolic dysfunction but also behavioral, emotional, and psychological components. People with obesity are more likely than the general population to develop depressive disorders, anxiety, social phobia, and disordered eating behaviors such as binge eating disorder. Recent studies have indicated a bidirectional relationship between obesity and depression, where each increases the risk of the other by approximately 1.5-fold across several cohorts \cite{luppino2010overweight}. In clinical settings, the coexistence of mental health comorbidities complicates obesity treatment. Depressive symptoms can reduce motivation for physical activity \cite{duina2025barriers}, while anxiety and emotional dysregulation may trigger disordered eating patterns \cite{trompeter2025prospective}. Overall, mental health challenges are associated with poorer treatment adherence, reduced response to standard therapies, and a lower quality of life among individuals with obesity \cite{gerardo2025depression}. Therefore, an effective obesity care model requires the incorporation of psychological, behavioral, and contextual dimensions, rather than an exclusive focus on weight and metabolism. It involves assessment, modeling, and personalization of interventions informed by patients’ psychological and behavioral characteristics, ensuring that mental health is incorporated as a fundamental aspect of obesity management.

However, our understanding of obesity patients with mental health comorbidities is constrained in practice by fragmented, sparse, and biased real-world clinical data. Many cohorts under-represent severe psychiatric phenotypes or lack consistent psychological assessments; missing modalities, inconsistent coding, and data sharing restrictions further limit joint modeling of metabolic and mental health dimensions. Synthetic data provide a viable means of generating artificial cohorts that preserve statistically realistic relationships among demographic, clinical, behavioral, and psychosocial variables, thereby allowing gap filling, feature interpolation, and investigation of rare or extreme phenotypes in silico while maintaining data privacy \cite{foraker2025understanding}. Furthermore, in the medical domain, synthetic cohorts enable hypothesis testing, model prototyping, and cross-site validation, including stress testing under low-frequency but clinically critical conditions that real datasets rarely capture \cite{gonzales2023synthetic}.

To address these limitations, we introduce a Multi-Agent System (MAS) designed to simulate realistic patients in obesity research. The MAS integrates multi-source evidence, including diagnostic trajectories, interventions, and comorbidity histories derived from medical claims, along with insights from clinical literature and behavioral, lifestyle, and psychosocial distributions obtained from population surveys such as the Centers for Disease Control and Prevention (CDC)’s National Health and Nutrition Examination Survey (NHANES) and the Behavioral Risk Factor Surveillance System (BRFSS). Furthermore, personality dimensions, six-factor model of personality encompassing Honesty–Humility, Emotionality, Extraversion, Agreeableness, Conscientiousness, and Openness to Experience (HEXACO) \cite{wang2025exploring}, Five reinforcement sensitivity theory (RST) \cite{krupic2016five}, and Temperament and Character Inventory (TCI) \cite{tomita2000factor}, are incorporated to assign each simulated patient a rich personality scale that modulates behavioral responses, emotional reactivity, and treatment adherence. This approach preserves the joint dependencies among metabolic, psychological, and personality-related features, enabling systematic exploration of personality traits relation with obesity–mental health trajectories and treatment outcomes. In doing so, this work bridges clinical, psychological, and behavioral data streams to simulate a more realistic obesity cohort. By uniting multi-agent modeling with generative synthesis, it lays the foundation for scalable, privacy-preserving, and psychologically informed research in obesity care.

\section{Related Work}

\subsection{Early Approaches to Simulate Realistic Patients}
Early approaches to simulate patients in healthcare used rule-based simulation engines. One prominent example, Synthea, employed modular disease and workflow scripts to simulate complete patient lifespans and produce structured electronic health records (EHRs) \cite{walonoski2018synthea}. Subsequent refinements introduced statistical calibration and multi-type event modeling to improve demographic and epidemiologic realism \cite{zhang2020ensuring,yan2021generating}. These rule-based systems remain valuable because they produce interpretable and schema-consistent records aligned with healthcare standards. However, their largely deterministic design limits the ability to capture high-dimensional temporal variability and subtle comorbidity patterns.

\subsection{Deep Generative Models for Patient Simulation}
To learn complex temporal and multimodal dependencies directly from real EHRs, researchers have adopted deep generative models. Early frameworks such as SynTEG and EVA introduced variational and adversarial autoencoders to generate longitudinal diagnosis sequences while quantifying fidelity and utility via Train-on-Synthetic-Test-on-Real (TSTR) evaluations \cite{zhang2021synteg,biswal2021eva}. EHR-M-GAN extended this idea to mixed continuous and categorical Intensive Care Unit (ICU) time series, demonstrating that synthetic data can improve predictive performance \cite{li23generating}. More recent transformer- and diffusion-based models improve realism by modeling irregular temporal gaps and heterogeneous data types \cite{yuan2023ehrdiff,karami2024synehrgy}. Together, these deep models advance fidelity and utility well beyond rule-based systems, yet they remain limited in representing clinical reasoning processes and multi-agent interactions across roles.

\subsection{LLMs, Multi-Agent Systems, and Clinical Narratives}
Recent advances in large language models (LLMs) have shifted synthetic-EHR research toward producing narrative-rich, context-aware records and agentic clinical simulations. Prompt-based approaches serialize structured EHRs as text for conditional generation \cite{wang2024promptehr}, while methods that couple structured encounters with visit-level or longitudinal notes improve contextual coherence and downstream text-classification utility \cite{sun2024collaborative}. Multi-role generative agents, in which clinician, patient, and coder LLMs collaborate to synthesize clinical documentation, mark a conceptual move toward agentic EHR generation \cite{velzen2025privacy}. However, many such systems remain text-centric, lacking full structured-code outputs, standards enforcement, or quantitative privacy and utility audits. Prior psychiatric-simulation frameworks (e.g., conversational agents for mental health) demonstrate realism in dialogue but not longitudinal data fidelity \cite{wang2024patient,lee2025psyche}. 

\section{Methods}

\subsection{Data}
The ultimate goal in this study is to simulate realistic patients in all dimensions including demographic, behavioral, and clinical diversity. As a result, we integrated multiple complementary datasets in the MAS framework. These sources jointly provide population-level distributions, longitudinal disease patterns, and case-level clinical detail. The integration followed a layered strategy: (i) National-level survey data for demographic and behavioral baselines, (ii) Medical claim dataset to capture longitudinal disease trajectories, (iii) Epidemiological data for probabilistic calibration, and (iv) Patient case reports for clinical and psychological realism. This multimodal design enables the multi-agent system (MAS) to simulate demographically representative and clinically coherent patients grounded in empirical data.

\subsubsection{NHANES}We used 12 survey cycles (1999–2023) from the CDC’s National Health and Nutrition Examination Surveys \cite{CDC_NCHS_NHANES_1999_2023}. These surveys integrates interviews, physical examinations, and laboratory tests to assess the health and nutrition of a nationally representative U.S. population. The combined dataset we extracted spans 21 health domains, covering demographic factors, anthropometric measurements, chronic and metabolic conditions, lifestyle behaviors, mental health indicators, clinical management, and specialized topics such as reproductive health and acculturation. Data across cycles were standardized and linked using the unique respondent identifier to create harmonized, participant-level records suitable for downstream synthesis.

\subsubsection{Medical Claims}We extracted 70,000 de-identified patients from the PurpleLab medical claims database (https://purplelab.com) who had at least one obesity-related claim within a 10-year lookback period. Each patient’s longitudinal record included demographics, diagnoses, treatment procedures, and medications, enabling reconstruction of detailed disease and treatment trajectories suitable for temporal modeling.

\subsubsection{Epidemiological Data}Population-level probabilities for obesity and selected psychological comorbidities were derived from multiple authoritative sources, including the CDC’s Behavioral Risk Factor Surveillance System Data \cite{BRFSS_CDC}, the World Obesity Federation \cite{WOF_US_ObesityData_2025}, and National Comorbidity Survey Replication \cite{kessler2004national}. These distributions defined sampling priors for demographic attributes (age, sex, race, education, occupation, income) and for targeted comorbidities of interest (binge-eating disorder, depression, anxiety disorder, and social phobia) ensuring that the simulated cohort reflects realistic obesity prevalence and comorbidity patterns observed in U.S. population data.

\subsubsection{Patient Case Reports}We collected patient case reports from the PubMed, a database of biomedical and life sciences literature. These reports, primarily published in peer-reviewed clinical and medical journals, contain detailed clinical narratives featuring diagnostic, therapeutic, and psychosocial information. Such content enables the targeted extraction of patient-level evidence relevant to specific conditions. Using this corpus, we curated case studies aligned with the diseases assigned to each simulated patient. The retrieved narratives provided empirical grounding for symptom patterns, behavioral characteristics, and treatment trajectories, which were subsequently utilized to enhance the realism and medically coherence of the simulated patient profiles.

\begin{figure*}[t!]
\centering
\includegraphics[width=\linewidth]{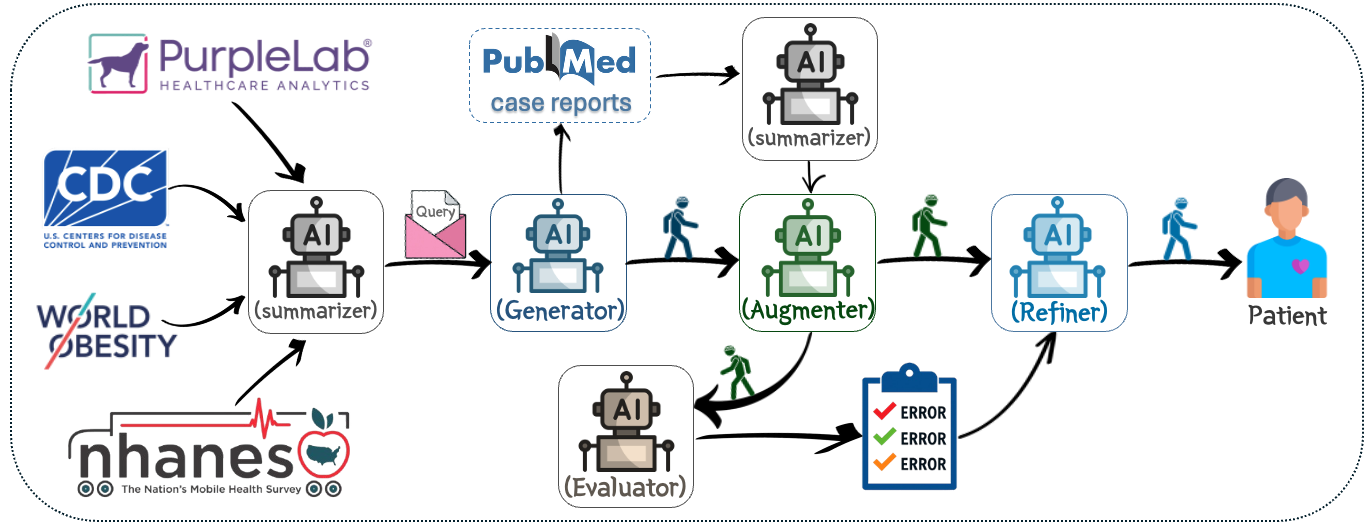}
\caption{Overview of the SynthAgent multi-agent framework for obesity patient simulation. The system integrates data from multiple empirical sources, including NHANES, medical claims, epidemiological datasets, and patient case reports. After data pre-processing, the summarizer agent condenses empirical evidence into structured inputs for the generator agent, which creates initial patient profiles. The augmenter agent enriches these profiles using literature-based clinical evidence, while the evaluator agent checks for logical, temporal, and behavioral inconsistencies. The refiner agent then resolves any issues to produce the final validated synthetic patient.}
\label{fig:pipeline}
\end{figure*}

\subsection{Generating Synthetic Patients Using MAS}
Our proposed methodology follows a multi-stage framework designed to simulate realistic and coherent patient records. The process begins with data pre-processing, where multiple sources, including NHANES surveys, medical claims, epidemiological datasets, and patient case reports, are standardized, harmonized, and integrated into unified patient-level records. After pre-processing, patient simulations are provided through a coordinated MAS composed of five different steps, each handled by a specialized agent. The summarizer agent condenses empirical data into structured inputs; the generator agent creates initial patient profiles; the augmenter agent enriches them with longitudinal details and evidence-based medical information; the evaluator agent audits for plausibility and consistency; and the refiner agent resolves the detected issues to produce the final validated profiles. Figure~\ref{fig:pipeline} illustrates the overall workflow.

\subsubsection{Summarizer Agent}
The summarizer agent initiates the MAS framework by constructing a foundational profile for each simulated patient. This process integrates epidemiological probabilities with real-world clinical data to ensure a realistic starting point. The agent's operation begins with the probabilistic generation of a demographic profile and the assignment of a specific condition from one of the designated comorbidity groups (i.e., obesity alone or obesity co-occurring with binge eating disorder, anxiety disorder, depression, or social phobia).

To ground this profile in authentic data, the agent draws from two distinct sources: a repository of 10 representative individual samples from NHANES data and a database of 20 longitudinal patient records from medical claims data. The agent then employs a matching algorithm to identify the single NHANES sample that most closely aligns with the generated demographic and comorbidity profile. Following this, it selects the top three clinical trajectories from the medical claims data that best match the combined attributes of the generated profile and the chosen NHANES sample.

Finally, the agent summarizes this information to produce a structured blueprint that guides the subsequent Generator Agent. This blueprint contains the patient's demographics, a specific BMI range (Class I, II, or III), the assigned comorbidity, a summarized version of the matched NHANES sample, and concise summaries of the three selected disease trajectories. This output serves as the foundational framework for simulating a complete and lifelike patient record.

\subsubsection{Generator Agent}
Once the summarizer agent defines each patient’s profile and potential health trajectory through summarized reference samples, the generator agent then transforms this blueprint into a complete, lifelike simulation. It transforms the structured summary into a full, clinically coherent simulated patient profile that reads like a realistic medical record. Guided by demographic and clinical parameters, such as age, BMI, and comorbidity patterns, the agent fills in every detail of the patient’s background, conditions, and experiences.

Each profile is generated through a controlled reasoning process that ensures accuracy and consistency. The agent constructs key sections that together define the patient’s complete health narrative: demographics and medical history, current conditions, symptoms, habits, and laboratory values. It also describes treatment plans, psychological traits, and behavioral tendencies that influence motivation and adherence. To make each record feel authentic, the agent includes a disease timeline that traces the patient’s journey over several years, showing how physical and psychological factors evolve over time.

\subsubsection{Augmenter Agent}Following the simulation of a base patient profile by the generator agent, the augmenter agent is tasked with enriching the simulation with evidence-based clinical and behavioral details to enhance its medical authenticity. The primary function of this agent is to revise the patient's record by integrating nuanced findings extracted directly from real-world medical literature.

The process begins by identifying all relevant disease keywords present in the generated patient profile. These keywords are used to query the PubMed database, retrieving 10 pertinent real-patient case reports for each term. This corpus of literature is then processed by the summarizer agent, which first filters the findings to exclude any cases that do not match the simulated patient's demographic profile, specifically age and gender. Subsequently, it analyzes the remaining case reports to synthesize a summary detailing how the documented conditions typically affect a patient's experiences, symptoms, behavioral patterns, and psychological states.

This synthesized clinical evidence is then passed to the augmenter agent, which concurrently receives the base patient profile from the generator agent. The augmenter agent meticulously integrates these insights into the profile, strategically refining sections such as symptoms, psychological scales, and the role-play Profile. For instance, if the synthesized evidence indicates a strong correlation between anxiety and sleep disturbances, this detail is woven into the synthetic patient’s behavioral narrative and symptomology. By grounding each record in evidence from real patient cases, this process bridges the gap between statistical realism and authentic human nuance, resulting in a final output of an augmented synthetic patient who not only aligns with clinical expectations but also exhibits the subtle emotional and behavioral complexities characteristic of real-world individuals.

\subsubsection{Evaluator Agent}
Once a simulated patient profile has been medically enriched, the evaluator agent acts as an automated clinical reviewer, ensuring that each record is logical, realistic, and internally consistent. It examines the complete profile through the lens of a virtual quality auditor, checking for errors or contradictions across multiple dimensions of a patient’s life and health.

The agent reviews demographic details to confirm that occupations, income levels, and insurance types make sense for a person’s age and background. It inspects medical information to verify that diseases, symptoms, and treatments align correctly and that no impossible combinations or temporal errors occur for example, a treatment appearing before a diagnosis. Psychological coherence is also assessed, ensuring that personality scores and behavioral traits match the patient’s reported habits and emotional patterns. Finally, the agent evaluates lifestyle realism, confirming that diet, activity, and sleep behaviors are consistent with the clinical picture.

After completing this review, the evaluator agent compiles a report that summarizes all detected issues and rates their severity as major, moderate, or minor. This report guides the next stage of the framework, helping the refiner agent to come up with the final simulated patient's profile.

\subsubsection{Refiner Agent}
The refiner agent serves as the final stage of the MAS framework, ensuring that every simulated patient's record is accurate, coherent, and ready for use. Acting like a clinical data expert, it reviews the report provided by the evaluator agent and makes precise adjustments to correct the simulated patient's profile. Rather than regenerating the entire record, the Refiner carefully edits only what is necessary to preserve the integrity of the patient’s story and maintain consistency across demographic, clinical, and psychological dimensions. For instance, fixing a timeline inconsistency, adjusting a mismatched diagnosis, or refining a behavioral detail. When no major issues are found, the agent performs a final quality sweep to detect subtle inconsistencies or overlooked details. The result is a polished, validated simulated patient's profile that faithfully represents realistic health and behavioral dynamics.

\subsection{Simulated Patient's Profile Evaluation}
We evaluated the MAS framework to assess the realism, coherence, and clinical quality of the generated synthetic patients. To fairly evaluate the impact of different core reasoning engines on the final simulation quality, we established four distinct MAS-configurations. The rationale for this design was to hold the input constant while varying the model responsible for the complex simulation tasks. To achieve this, the summarizer agent, powered by Gemini 2.0 Flash, was held constant across all experiments. This agent produced 30 standardized foundational profiles, which served as the identical inputs for all four pipeline configurations, ensuring a controlled comparison.

The primary variable differentiating the four MAS-configurations was the LLM used as the reasoning engine for the entire core agent suite: the generator, augmenter, evaluator, and refiner. We tested four leading LLM engines in this role: GPT-5 \cite{openai2025gpt5}, Gemini 2.5 Pro \cite{comanici2025gemini}, Claude 4.5 Sonnet \cite{anthropic2025claudesonnet4_5}, and DeepSeek-R1 \cite{deepseek2025r1}. 

The goal was to have high quality while the cohort is also diverse. Therefore, evaluation combined two complementary analyses: (i) Automated quality assessment, which used an LLM-as-a-Judge framework as an examiner to score based on the coherence, realism, and completeness, and (ii) Latent diversity visualization, which analyzed embedding-based variation among the simulated patients.

For the automated quality assessment, we implemented an LLM-as-a-Judge framework, leveraging GPT-4o \cite{openai2024gpt4o} as an independent evaluator to ensure consistent, human-like judgment across all 120 simulated patient profiles. Each profile was systematically scored using a detailed schema covering all dimensions to evaluate the reflection of both medical accuracy and narrative coherence: (1) demographics, (2) medical history, (3) current conditions, (4) symptoms, (5) habits, (6) laboratory values, (7) treatments, (8) psychological scales, (9) role-play profile, and (10) disease timeline.

The scoring followed a deductive scheme, starting from a perfect score of 100, with points deducted for specific errors identified across three core criteria: Information Sufficiency, which determined if the details were comprehensive and well-proportioned; Logical Consistency, which verified internal coherence between different sections of the profile (e.g., ensuring psychological conditions aligned with behavioral patterns). and Medical Plausibility, which confirmed that clinical details were realistic (e.g., treatments were appropriate for the stated conditions and symptoms). This rigorous process produced both individual dimension scores and a final overall quality score for each simulated patient, allowing for granular and holistic evaluation.

To compare model performance, all scores were aggregated and measured using descriptive statistics (mean, median, standard deviation, and range). Pairwise model comparisons were conducted using Student's t-tests to determine whether performance were statistically different. We considered results statistically significant when $p < 0.05$.

To assess the semantic diversity and distinctiveness of the simulated cohort, we performed a comprehensive embedding-based analysis on the simulated patient cohort that was generated by different configuration of MAS framework (i.e., four LLM engines). Jina Embeddings v4~\cite{gunther2025jina} was employed to transform patient records into high-dimensional semantic representations (768 dimensions). Demographic information was excluded from the cohort diversity analysis to ensure that diversity measures reflected differences in clinical and behavioral content rather than demographic variation, allowing a fair comparison of overall patient record diversity. Next, we computed within MAS-configuration diversity scores by measuring the mean Euclidean distance from each patient embedding to the centroid of its respective MAS-configuration's cohort. Between MAS-configuration similarity was assessed using cosine similarity and pairwise distances. For visualization, we applied t-SNE~\cite{maaten2008visualizing} to project the embeddings into a 2D space, representing patients as segmented circles to indicate their comorbidity profiles (obesity with depression, anxiety disorder, social phobia, binge eating disorder, or obesity alone).

This comprehensive evaluation provided an objective, scalable method for measuring the coherence and clinical fidelity of simulated patient data, while accounting for cohort diversity, thereby enabling quantitative comparisons of different language model backbones within the MAS framework.

\begin{figure*}[t!]
    \centering
    \includegraphics[width=\textwidth]{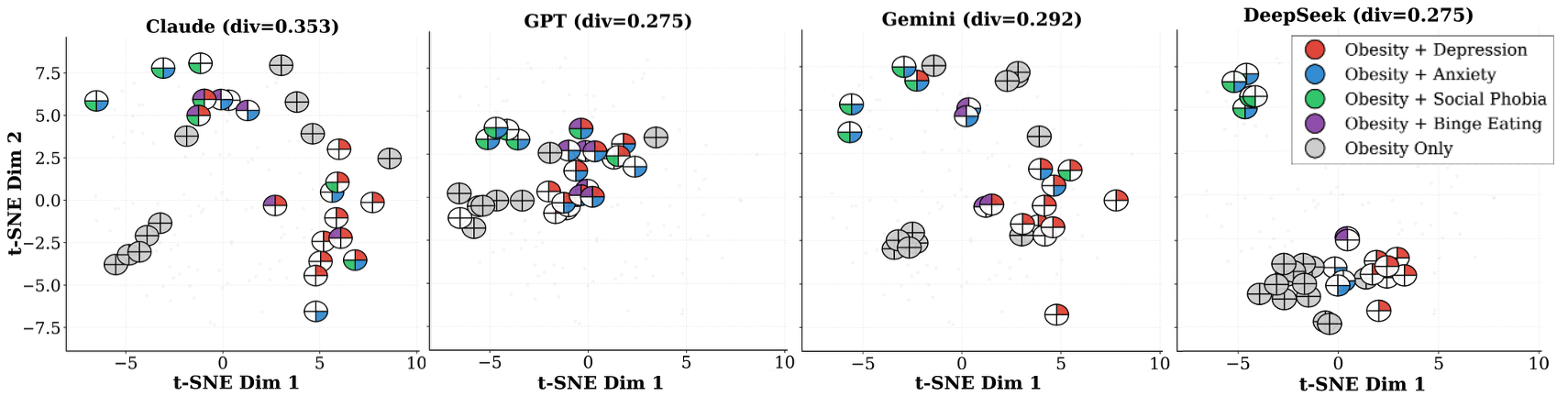}
    \caption{t-SNE visualization of clinical core embeddings across four LLMs. Segmented circles indicate comorbidity patterns (red=depression, blue=anxiety, green=social phobia, purple=binge eating, gray=obesity only).}
    \label{fig:categories_clinical_core}
\end{figure*}

\section{Results}
\subsection{Qualitative Assessment}We evaluated four LLMs (GPT-5, Claude 4.5 Sonnet, Gemini 2.5 Pro, and DeepSeek-R1) as engines for patient simulation in the MAS framework. From 120 simulated patients (30 per engine), we found that GPT-5 achieved the highest mean quality score (76.27~$\pm$~2.84), followed closely by Claude~4.5~Sonnet (76.17~$\pm$~3.90), the engine that generated the single highest-rated synthetic patient (score = 82; Table~\ref{tab:overall_quality}). The performance of these two leading models show similar performance of patient simulation ($p = 0.91$). In contrast, Gemini~2.5~Pro (71.2~$\pm$~3.0) and DeepSeek-R1 (67.8~$\pm$~3.3) exhibited significantly lower performance (Table~\ref{tab:pairwise_tests}).

\begin{table}[ht!]
\centering
\caption{Overall performance evaluation of LLM engines.}
\label{tab:overall_quality}
\small
\begin{tabular}{lcccc}
\toprule
\textbf{LLM} & \textbf{Mean $\pm$ SD} & \textbf{Min} & \textbf{Max} & \textbf{Median} \\
\midrule
GPT-5     & 76.27~$\pm$~2.84 & 72 & 81 & 76.0 \\
Claude    & 76.17~$\pm$~3.90 & 69 & 82 & 76.0 \\
Gemini    & 71.17~$\pm$~2.96 & 65 & 77 & 71.0 \\
DeepSeek  & 67.83~$\pm$~3.26 & 63 & 76 & 68.0 \\
\bottomrule
\end{tabular}
\end{table}

\begin{table}[ht!]
\centering
\caption{Pairwise comparison of LLM engines.}
\label{tab:pairwise_tests}
\small
\begin{tabular}{lcc}
\toprule
\textbf{Comparison} & \textbf{p-value} & \textbf{Interpretation} \\
\midrule
Claude vs GPT-5    & 0.910 & Equivalent \\
Claude vs Gemini   & $<$0.001 & Claude $\gg$ Gemini \\
Claude vs DeepSeek & $<$0.001 & Claude $\gg$ DeepSeek \\
GPT-5 vs Gemini    & $<$0.001 & GPT-5 $\gg$ Gemini \\
GPT-5 vs DeepSeek  & $<$0.001 & GPT-5 $\gg$ DeepSeek \\
Gemini vs DeepSeek & 0.0001 & Gemini $\gg$ DeepSeek \\
\bottomrule
\end{tabular}
\end{table}

The dimension-level evaluation further supported our previous observations. While GPT-5 and Claude 4.5 Sonnet generally outperformed Gemini 2.5 Pro and DeepSeek-R1 at this granular level, DeepSeek-R1 notably excelled in simulating the patient's current condition. GPT-5 demonstrated superior performance in technical and structural categories, including medical history, treatments, role- play profile, and disease timeline. Conversely, Claude 4.5 Sonnet excelled in dimensions such as symptoms, habits, labs, and psychological scales. These findings suggest that GPT-5's strengths lie in structural and temporal coherence, while Claude 4.5 Sonnet provides richer psychological realism. Detailed dimension-level comparisons are available in Table~\ref{tab:dimension_summary}.

\begin{table}[ht!]
\centering
\caption{Dimension-level quality scores of LLM engines.}
\label{tab:dimension_summary}
\small
\begin{tabular}{lcccc}
\toprule
\textbf{Dimension} & \textbf{Claude} & \textbf{GPT-5} & \textbf{Gemini} & \textbf{DeepSeek} \\
\midrule
Medical History     & 99.6 & \textbf{100.0} & 95.8 & 95.8 \\
Current Conditions  & 84.9 & 84.4 & 84.4 & \textbf{86.2} \\
Symptoms            & \textbf{79.1} & 76.2 & 73.6 & 70.7 \\
Habits              & \textbf{61.7} & 53.8 & 52.5 & 43.3 \\
Labs                & \textbf{61.1} & 52.3 & 52.3 & 59.0 \\
Treatments          & 70.9 & \textbf{80.0} & 73.3 & 65.1 \\
Psych Scales        & \textbf{71.3} & 67.7 & 53.3 & 49.3 \\
Role Play           & 57.0 & \textbf{60.7} & 50.3 & 48.7 \\
Timeline            & 81.0 & \textbf{82.3} & 77.2 & 69.5 \\
\bottomrule
\end{tabular}
\end{table}

\subsection{Cohort Diversity Analysis} 
Our embedding-based analysis revealed significant differences in the semantic diversity of the simulated patient cohort generated by the four different MAS configurations. Claude 4.5 Sonnet core engine produced the most varied patient profiles, achieving the highest diversity score (div~=~0.353). Conversely, DeepSeek-R1 and GPT-5 generated more semantically homogenous cohorts, evidenced by lower diversity scores (div~=~0.275 for both).

These quantitative findings are visually corroborated by the 2D t-SNE projection of the patient embeddings. As shown in  Figure~\ref{fig:categories_clinical_core}, each circle represents a patient simulated with obesity, and colored segments indicate specific comorbidities such as depression or anxiety. The arrangement of these circles in the 2D space reflects their semantic similarity. We observed a dense clustering of patients generated by DeepSeek-R1 and Gemini 2.5 Pro, particularly for single-condition profiles (obesity only), indicating a high degree of semantic similarity within their respective cohorts. In contrast, the patient embeddings from GPT-5, and most notably Claude 4.5 Sonnet, are distributed more broadly across the semantic space. The dispersion for the Claude 4.5 Sonnet cohort is particularly pronounced, demonstrating clear separation and rich variability even among patients who share identical comorbidity profiles. This confirms its superior capacity for generating diverse and distinct clinical narratives in this simulation.

\subsection{Case Study}
\begin{figure}[ht!]
\centering
\includegraphics[width=0.98\columnwidth, height=0.9\textheight, keepaspectratio]{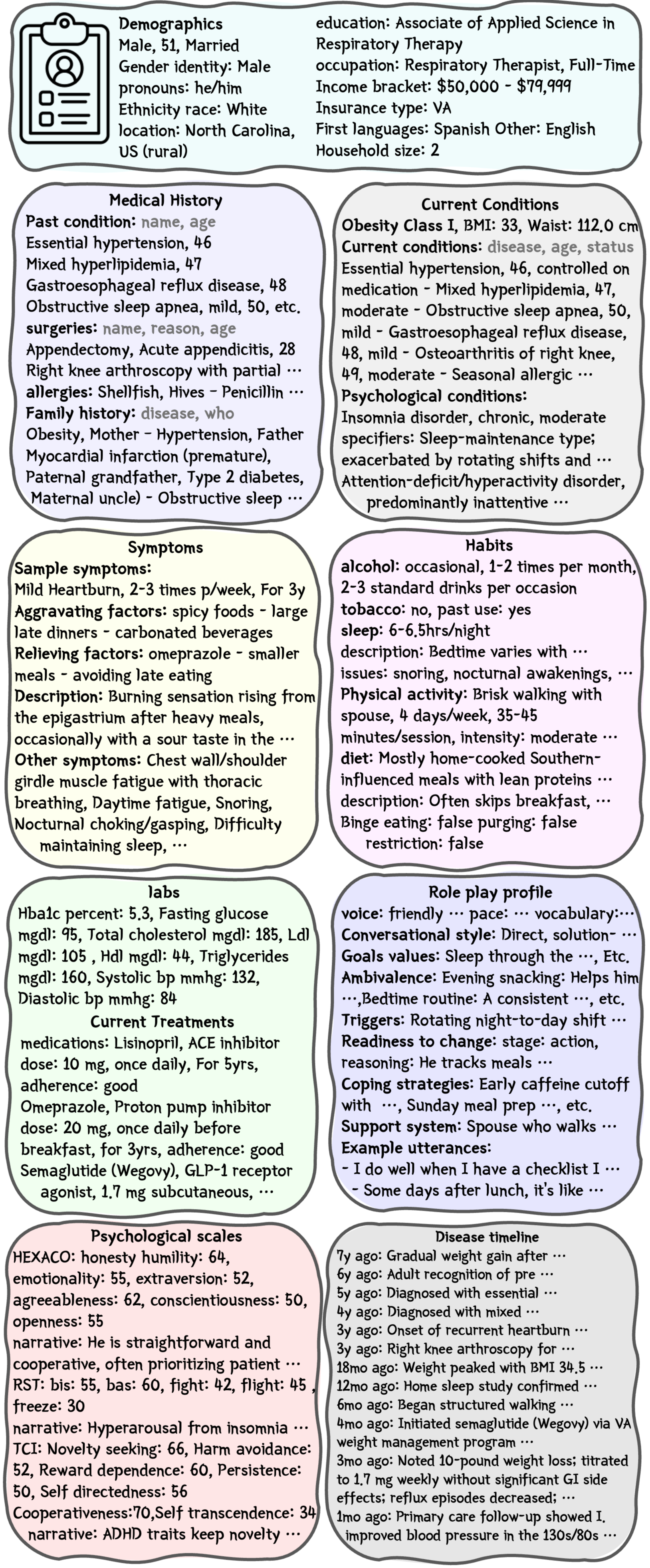}
\caption{A sample simulated patient generated by the proposed Multi-Agent System (MAS) using the GPT-5 LLM engine. Several dimensions are truncated to fit within the available space, and ellipses (…) indicate that the information continues. The patient’s overall score is 81.}
\label{fig:samplepatient}
\end{figure}

We provide details of one simulated patient as a case study to demonstrate our approach (Figure~\ref{fig:samplepatient}). In our simulation, the patient profile shows a 51-year-old married male living in North Carolina, US, with Class I obesity (BMI 33) alongside comorbid conditions hypertension, hyperlipidemia, obstructive sleep apnea, Gastroesophageal reflux disease and osteoarthritis, and documented psychological/behavioural features such as insomnia. The comprehensive lab values (HbA1c 5.3\%, fasting glucose 95 mg/dL, total cholesterol 185 mg/dL, LDL 105 mg/dL, HDL 44 mg/dL, triglycerides 160 mg/dL) align with known metabolic signatures of overweight or obesity class I patient and comorbid mental/behavioural disorders. For example, research in patients with stable mental disorders found a positive correlation of BMI with triglycerides and LDL-cholesterol in such populations \cite{li2022metabolic}. Furthermore, the documented psychological features, such as attention-deficit/hyperactivity disorder traits, sleep-disturbance and behavioural habits, reflect the well-established link between obesity and mental health burdens (i.e. people living with obesity have increased odds of depression, anxiety or eating-disorders) \cite{segal2024psychological}. The richness of this profile, combining demographic, medical-history, symptoms, habits, psychological scales and disease timeline demonstrates that the proposed MAS framework has successfully produced a structurally coherent and clinically credible simulated patient. The overall score of 81 further confirms that the profile meets a high-quality threshold for multi-dimensional patient simulation.

\section{Discussion}
\subsection{Our Contributions}
The emergence of multi-agent patient simulation framework represents a transformative step toward overcoming data scarcity and fragmentation in healthcare research. Traditional clinical datasets often lack the psychological, behavioral, and contextual richness required to understand complex, comorbid conditions such as obesity with mental disorders. By integrating diverse data sources (i.e., medical claims, NHANES surveys, and epidemiological data), SynthAgent demonstrates how coordinated multi-agent architectures can collectively emulate the multifactorial nature of real patients. Each agent contributes a specialized capability creating an adaptive simulation environment that mirrors real-world variability while maintaining privacy. This approach not only supports reproducible and ethically safe experimentation but also enables hypothesis testing under scenarios rarely observed in clinical datasets, such as treatment lapses triggered by emotional or cognitive stressors.

Beyond its application to obesity, SynthAgent highlights the potential of multi-agent frameworks to advance patient-centered healthcare modeling. By capturing decision-making processes, motivation shifts, and psychosocial adaptation over time, these systems provide a lens into the lived experience of patients, bridging clinical outcomes with behavioral dynamics. Such simulated cohorts can inform digital therapeutics, precision behavioral interventions, and AI-driven decision support tools by offering an interpretable, testable representation of patient diversity. As healthcare AI moves toward more human-centric models, multi-agent patient simulation frameworks like SynthAgent will play a pivotal role in enabling cross-disciplinary research, training next-generation clinical AI systems, and reimagining how we understand patients as dynamic, adaptive systems rather than static data points.

\subsection{Comparative Performance of LLM Engines in Patient Simulation}
The results demonstrate that while the overall quality of synthetic patients generated by GPT-5 and Claude 4.5 Sonnet is statistically equivalent, the underlying models exhibit distinct and complementary strengths. GPT-5 excels in producing structurally and temporally coherent patient narratives, evidenced by its superior performance in dimensions such as Medical History, Treatments, and Disease Timeline. This suggests its suitability for simulations where longitudinal consistency and historical accuracy are paramount. Conversely, Claude 4.5 Sonnet distinguishes itself by generating semantically diverse patient cohorts with greater psychological realism, outperforming other models in dimensions like Symptoms, Habits, and Psychological Scales. Crucially, the patient diversity for Claude 4.5 Sonnet was significantly higher than for GPT-5; this increased variety can compensate for any perceived minor differences in quality scores, positioning it as a potentially superior engine for the proposed MAS for patient simulations. The significantly lower performance of Gemini 2.5 Pro and DeepSeek-R1 suggests they are less optimal for this task, although DeepSeek-R1 showed a specific aptitude for articulating the patient's immediate condition. Ultimately, while GPT-5 is preferable for structural fidelity, Claude 4.5 Sonnet's proven strength in generating a diverse and psychologically rich cohort makes it an exceptionally strong choice. These findings also suggest the potential for future hybrid systems that leverage the unique strengths of different models to create even more robust and realistic simulated patient data. 

\subsection{Limitation}While SynthAgent demonstrates the potential of a MAS for simulating medically and psychologically enriched patients, several limitations remain. First, the framework’s performance depends on the quality and representativeness of input data such as medical claims and population surveys, which may introduce inherent biases or underrepresent rare psychiatric phenotypes. Second, although the MAS framework enables modular coordination among agents, inter-agent reasoning and conflict resolution remain partially rule-based, limiting emergent behavioral complexity. Third, psychological and behavioral modeling relies on proxy variables such as personality traits and survey-derived distributions, which cannot fully capture real-world variability or sociocultural nuance. Finally, the current evaluation focuses on internal quality and semantic diversity rather than external clinical validation. Future work should integrate clinician-in-the-loop assessment, reinforcement-driven agent adaptation, and cross-domain validation to ensure that simulated patients more faithfully reflect real patient dynamics and support translational healthcare research.

\section{Conclusion}
In this study, we introduced SynthAgent, a multi-agent simulation framework capable of generating medically, psychologically, and behaviorally enriched patient profiles. Its capabilities were demonstrated using obesity with comorbid mental disorders as a representative condition. Unlike single-model generation approaches, SynthAgent employs coordinated agents for data fusion, behavioral simulation, and quality evaluation, integrating medical claims, clinical evidence, and psychosocial distributions to produce realistic patient trajectories. Our findings highlight the potential of multi-agent synthetic patient generation to fill data gaps, enable ethically safe experimentation, and capture complex patterns of treatment adherence, emotion regulation, and lifestyle adaptation. By combining medical realism with psychological depth, SynthAgent establishes a scalable foundation for advancing patient-centered modeling and offers broad applicability for understanding diverse patient populations, decision-making processes, and health management behaviors across conditions.

% --- SORTED BIBLIOGRAPHY FOR CORRECT NUMBERING [1], [2], [3] ---

\end{document}